\documentclass[10pt,conference,a4paper]{IEEEtran}
\usepackage{amssymb}
\usepackage{graphicx}
\usepackage{multirow}
\usepackage{longtable}
\usepackage{rotating}
\usepackage{color}
\usepackage{hyperref}
\graphicspath{{figures/}}

\ifCLASSINFOpdf
\else
\fi
\begin{document}
%
\title{Infrared and Visible Image Fusion using a Deep Learning Framework}

\author{\IEEEauthorblockN{Hui~Li}
\IEEEauthorblockA{Jiangsu Provincial Engineering \\Laboratory of Pattern Recognition and \\Computational Intelligence,\\
Jiangnan University, \\
Wuxi, China,  214122\\
Email: hui\_li\_jnu@163.com}
\and
\IEEEauthorblockN{Xiao-Jun~Wu$^*$}
\IEEEauthorblockA{Jiangsu Provincial Engineering \\Laboratory of Pattern Recognition and \\Computational Intelligence,\\
Jiangnan University, \\
Wuxi, China,  214122\\
Email: xiaojun\_wu\_jnu@163.com}
\and
\IEEEauthorblockN{Josef Kittler}
\IEEEauthorblockA{\\ CVSSP \\
University of Surrey, \\
GU2 7XH, Guildford, UK \\\\
Email: j.kittler@surrey.ac.uk}}

\maketitle

\begin{abstract}
In recent years, deep learning has become a very active research tool which is used in many image processing fields. In this paper, we propose an effective image fusion method using a deep learning framework to generate a single image which contains all the features from infrared and visible images. First, the source images are decomposed into base parts and detail content. Then the base parts are fused by weighted-averaging. For the detail content, we use a deep learning network to extract multi-layer features. Using these features, we use $l_1$-norm and weighted-average strategy to generate several candidates of the fused detail content. Once we get these candidates, the max selection strategy is used to get the final fused detail content. Finally, the fused image will be reconstructed by combining the fused base part and the detail content. The experimental results demonstrate that our proposed method achieves state-of-the-art performance in both objective assessment and visual quality. The Code of our fusion method is available at \emph{\url{https://github.com/hli1221/imagefusion\_deeplearning}}.
\end{abstract}


\IEEEpeerreviewmaketitle

\section{Introduction}
The fusion of infrared and visible imaging is an important and frequently occuring problem. Recently, many fusion methods have been proposed to combine the features present in infrared and visible images into a single image\cite{1}. These state-of-the-art methods are widely used in many applications, like image pre-processing, target recognition and image classification. 

The key problem of image fusion is how to extract salient features from the source images and how to combine them to generate the fused image.

For decades, many signal processing methods have been applied in the image fusion field to extract image features, such as discrete wavelet transform(DWT)\cite{2}, contourlet transform\cite{3}, shift-invariant shearlet transform\cite{4} and quaternion wavelet transform\cite{5} etc. For the infrared and visible image fusion task, Bavirisetti et al. \cite{6} proposed a two-scale decomposition and saliency detection-based fusion method, where by the mean and median filter are used to extract the base layers and detail layers. Then visual saliency is used to obtain weight maps. Finally, the fused image is obtained by combining these three parts.

Besides the above methods, the role of sparse representation(SR) and low-rank representation has also attracted great attention. Zong et al.\cite{7} proposed a medical image fusion method based on SR, in which, the Histogram of Oriented Gradients(HOG) features are used to classify the image patches and learn several sub-dictionaries. The $l_1$-norm and the max selection strategy are used to reconstruct the fused image. In addition, there are many methods based on combining SR and other tools for image fusion, such as pulse coupled neural network(PCNN)\cite{8} and shearlet transform\cite{9}. In the sparse domain, the joint sparse representation\cite{10} and cosparse representation\cite{11} were also applied in the image fusion field. In the low-rank category, Li et al.\cite{12} proposed a low-rank representation(LRR)-based fusion method. They use LRR instead of SR to extract features, then $l_1$-norm and the max selection strategy are used to reconstruct the fused image.

With the rise of deep learning, deep features of the source images which are also a kind of saliency features are used to reconstruct the fused image. In \cite{13}, Yu Liu et al. proposed a fusion method based on convolutional sparse representation(CSR). The CSR is different from deep learning methods, but the features extracted by CSR are still deep features. In their method, the authors employ CSR to extract multi-layer features, and then use these features to generate the fused image. In addition, Yu Liu et al.\cite{14} also proposed a convolutional neural network(CNN)-based fusion method. They use image patches which contain different blur versions of the input image to train the network and use it to get a decision map. Finally, the fused image is obtained by using the decision map and the source images. Although the deep learning-based methods achieve better performance, these methods still have many drawbacks: 1) The method in \cite{14} is only suitable for multi-focus image fusion; 2) These methods just use the result which is calculated by the last layers and a lot of useful information which is obtained by the middle layers will be lost. The information loss tends to get worse when the network is deeper.

In this paper, we propose a novel and effective fusion method based on a deep learning framework for infrared and visible image fusion. The source images are decomposed into base parts and detail content by the image decomposition approach in \cite{15}. We use a weighted-averaging strategy to obtain the fused base part. To extract the detail, first, we use deep learning network to compute multi-layer features so as to preserve as much information as possible. For the features at each layer, we use soft-max operator to obtain weight maps and a candidate fused detail content will be obtained. Applying the same operation at multiple layers, we will get several candidates for the fused detail content. The final fused detail image is generated by the max selection strategy. The final fused image is reconstructed by fusing the base part with the detail content.

This paper is structured as follows. In Section\ref{sec:style}, the image style transfer using deep learning framework will be presented. In Section\ref{sec:proposed}, the proposed deep learning based image fusion method is introducted in detail. The experimental results are shown in Section\ref{sec:experiment}. Finally, Section\ref{sec:con} draws the paper to conclusion.

\section{Image style transfer using deep learning framework}
\label{sec:style}
As we all know, deep learning achieves the state-of-the-art performance in many image processing tasks, such as image classification. In addition, deep learning also can be a useful tool for extracting image fearures which contain different information at each layer. Different applications of deep learning received a lot of attention in the last two years. Hence, we believe deep learning can also be applied to the image fusion task.

In CVPR 2016, Gatys et al.\cite{16} proposed an image style transfer method based on CNN. They use VGG-network\cite{17} to extract deep features at diffierent layers from the ``content'' image, ``style'' image and a generated image, respectively. The difference of deep features extracted from the generated image and source images is mimimised by iteration. The generated image will contain the main object from the ``content'' image and texture features from the ``style'' image. Although this method can obtain good stylized image, its speed is extremly slow even when using GPU.

Due to these drawbacks, in ECCV 2016, Justin Johnson et al.\cite{18} proposed a feed-forward network to solve the optimization problem formulated in \cite{16} in real time. But in this method, each network is tied to a fixed style. To solve this problem, in ICCV 2017, Xun Huang et al.\cite{19} used VGG-network and adaptive instance normalization to construct a new style transfer framework. In this framework, the stylized image can be of arbitrary style and the method is nearly three orders of magnitude faster than \cite{16}.

These methods have one thing in common. They all use multi-layer network fearures as a constraint condition. Inspired by them, multi-layer deep features are extracted by a VGG-network in our fusion method. We use the fixed VGG-19\cite{17} which is trained on ImageNet to extract the features. The detail of our proposed fusion method will be introduced in the next section.

\section{The Proposed Fusion Method}
\label{sec:proposed}
The fusion processing of base parts and detail content is introduced in the next subsections.

Suppose that there are $K$ preregistered source images, in our paper, we choose $K=2$, but the fusion strategy is the same for $K>2$. The source images will be denoted as $I_k$, $k\in{\{1,2\}}$.

Compared with other image decomposition methods, like wavelet decomposition and latent low-rank decomposition, the optimization method\cite{15} is more effective and can save time. So in our paper, we use this method to decompose the source images.

For each source image $I_k$, the base parts $I_k^b$ and detail content $I_k^d$ are obtained separated by \cite{15}. The base parts are obtained by solving this optimization problem:
\begin{eqnarray}\label{Eq1}
  	I_k^b = \arg\min_{I_k^b}||I_k-I_k^b||_F^2+\lambda(||g_x*I_k^b||_F^2+||g_y*I_k^b||_F^2)
\end{eqnarray}

\noindent where $g_x=[-1\quad1]$ and $g_y=[-1\quad1]^T$ are the horizontal and vertical gradient operators, respectively. The parameter $\lambda$ is set to 5 in our paper.

After we get the base parts $I_k^b$, the detail content is obtained by Eq.\ref{Eq2},
\begin{eqnarray}\label{Eq2}
  	I_k^d = I - I_k^b
\end{eqnarray}

\begin{figure}[!ht]
\centering
\includegraphics[width=\linewidth]{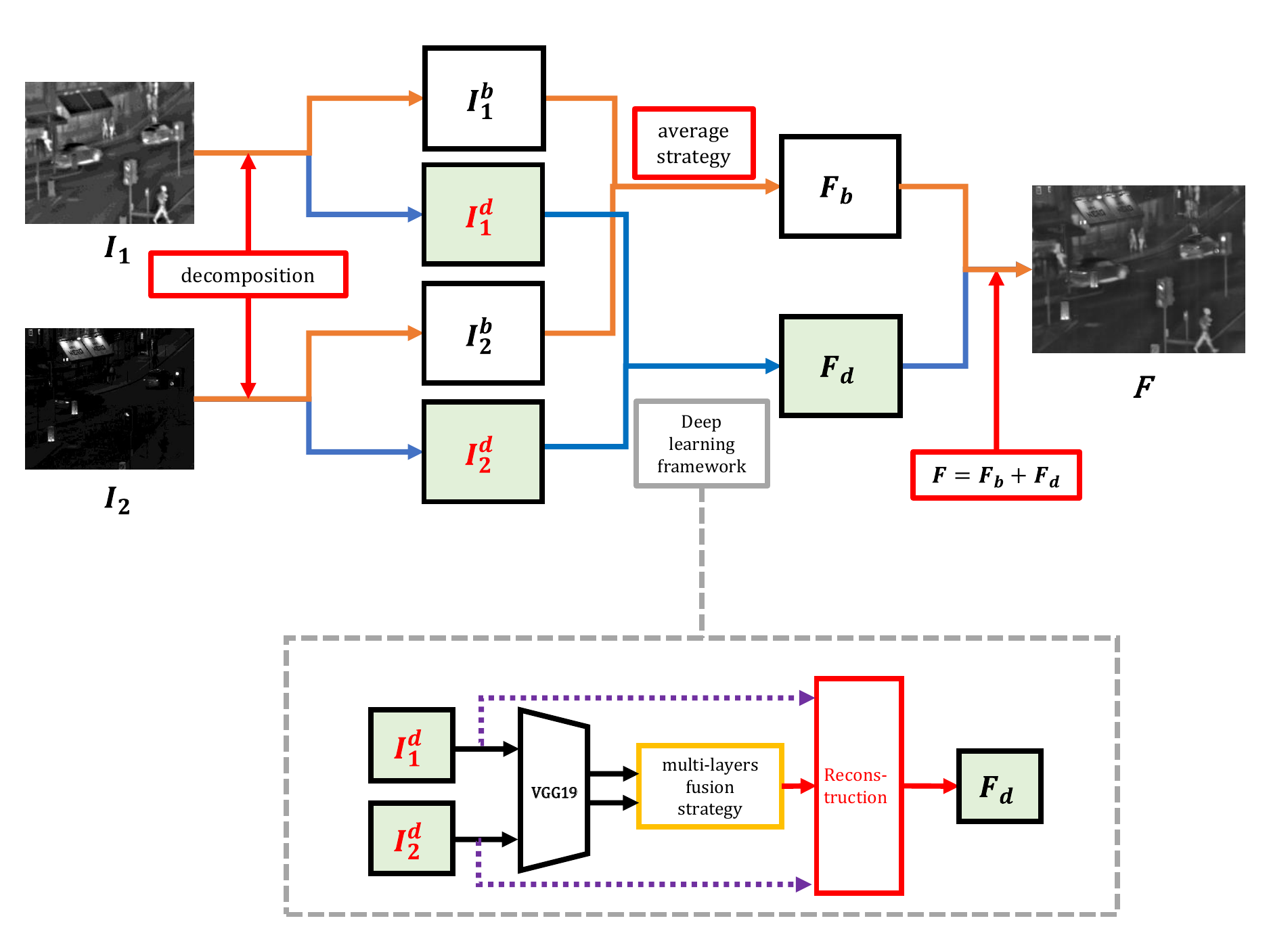}
\caption{The framework of proposed method.}
\label{fig1}
\end{figure}

The framework of the proposed fusion method is shown in Fig.\ref{fig1}.

As shown in Fig.1, the source images are denoted as $I_1$ and $I_2$. Firstly, the base part $I_k^b$ and the detail content $I_k^d$ for each source image are obtained by solving Eq.(1) and Eq.(2), where $k\in{\{1,2\}}$. Then the base parts are fused by weighted-averaging strategy and the detail content is reconstructed by our deep learning framework. Finally, the fused image $F$ will be reconstructed by adding the fused base part $F_b$ and detail content $F_d$.

\subsection{Fusion of base parts}
The base parts which are extracted from the source images contain the common features and redundant information. In our paper, we choose the weighted-averaging strategy to fuse these base parts. The fused base part is calculated by Eq.\ref{Eq3},
\begin{eqnarray}\label{Eq3}
  	F_b(x,y) = \alpha_{1}I_1^b(x,y)+\alpha_{2}I_2^b(x,y)
\end{eqnarray}

\noindent where $(x,y)$ denotes the corresponding position of the image intensity in $I_1^b$, $I_2^b$ and $F_b$. $\alpha_1$ and $\alpha_2$ indicate the weight values for pixel in $I_1^b$ and $I_2^b$, respectively. To preserve the common features and reduce the redundant information, in this paper, we choose $\alpha_1=0.5$ and $\alpha_2=0.5$.

\subsection{The fusion of the detail content}

\begin{figure}[!ht]
\centering
\includegraphics[width=\linewidth]{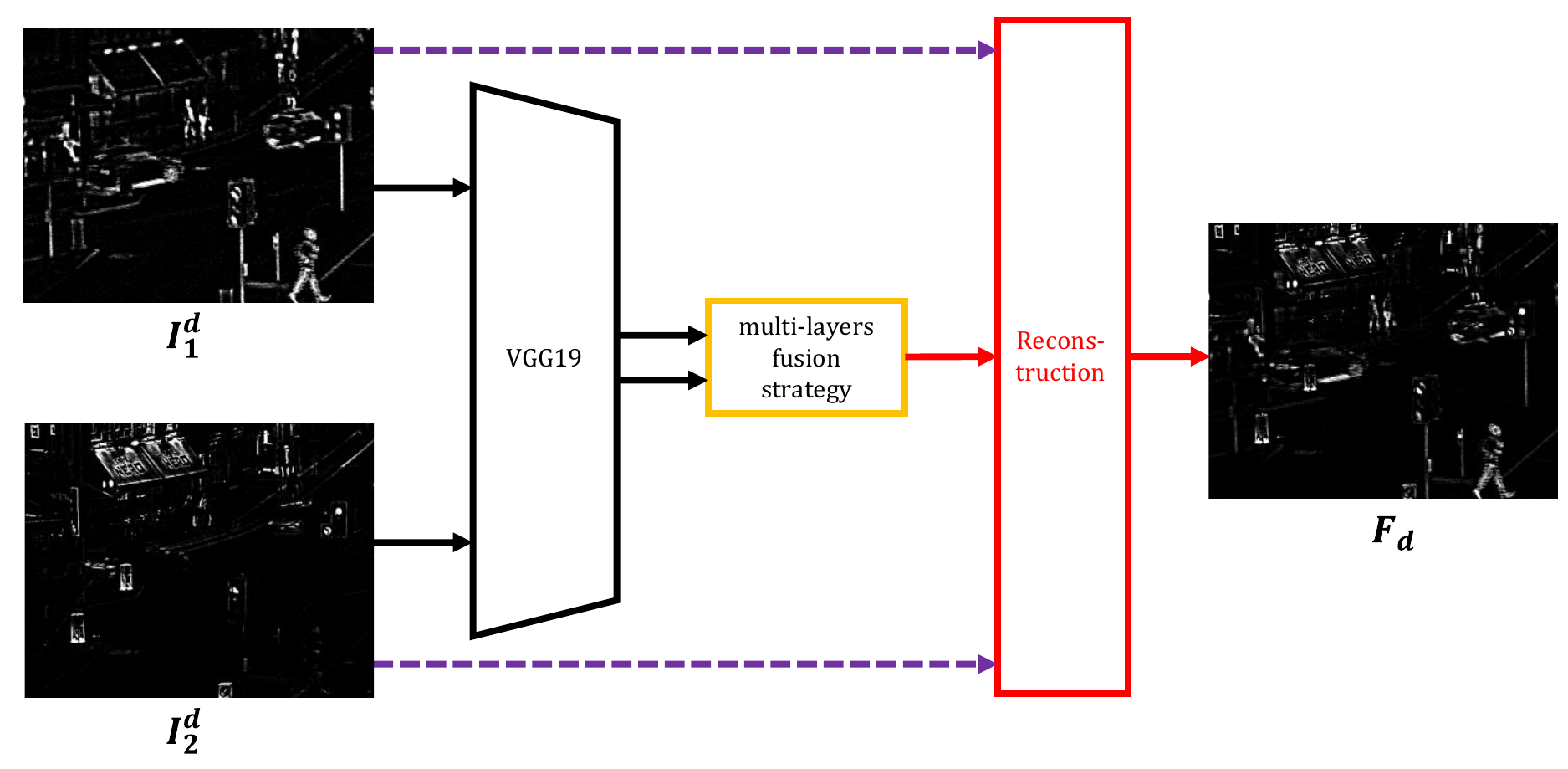}
\caption{The procedure of detail content fusion.}
\label{fig2}
\end{figure}

For the detail content $I_1^d$ and $I_2^d$, we propose a novel fusion strategy which uses deep learning method(VGG-network) to extract deep features. This procedure is shown in Fig.\ref{fig2}.

In Fig.\ref{fig2}, we use VGG-19 to extract deep features. Then the weight maps are obtained by a multi-layer fusion strategy. Finally, the fused detail content is reconstructed by these weight maps and the detail content.

Now, we introduce our multi-layer fusion strategy in detail.

Consider detail content $I_k^d$. $\phi_k^{i,m}$ indicates the feature maps of $k$-th detail content extracted by the $i$-th layer and m is the channel number of the $i$-th layer, $m\in{\{1,2,\cdots,M\}}$,$M=64\times2^{i-1}$,
\begin{eqnarray}\label{Eq4}
  	\phi_k^{i,m} = \Phi_i(I_k^d)
\end{eqnarray}

\noindent where each $\Phi_{i}(\cdot)$ denotes a layer in the VGG-network and $i\in{\{1,2,3,4\}}$ represents the $relu\_1\_1$, $relu\_2\_1$, $relu\_3\_1$ and $relu\_4\_1$, respectively.

\begin{figure}[!ht]
\centering
\includegraphics[width=\linewidth]{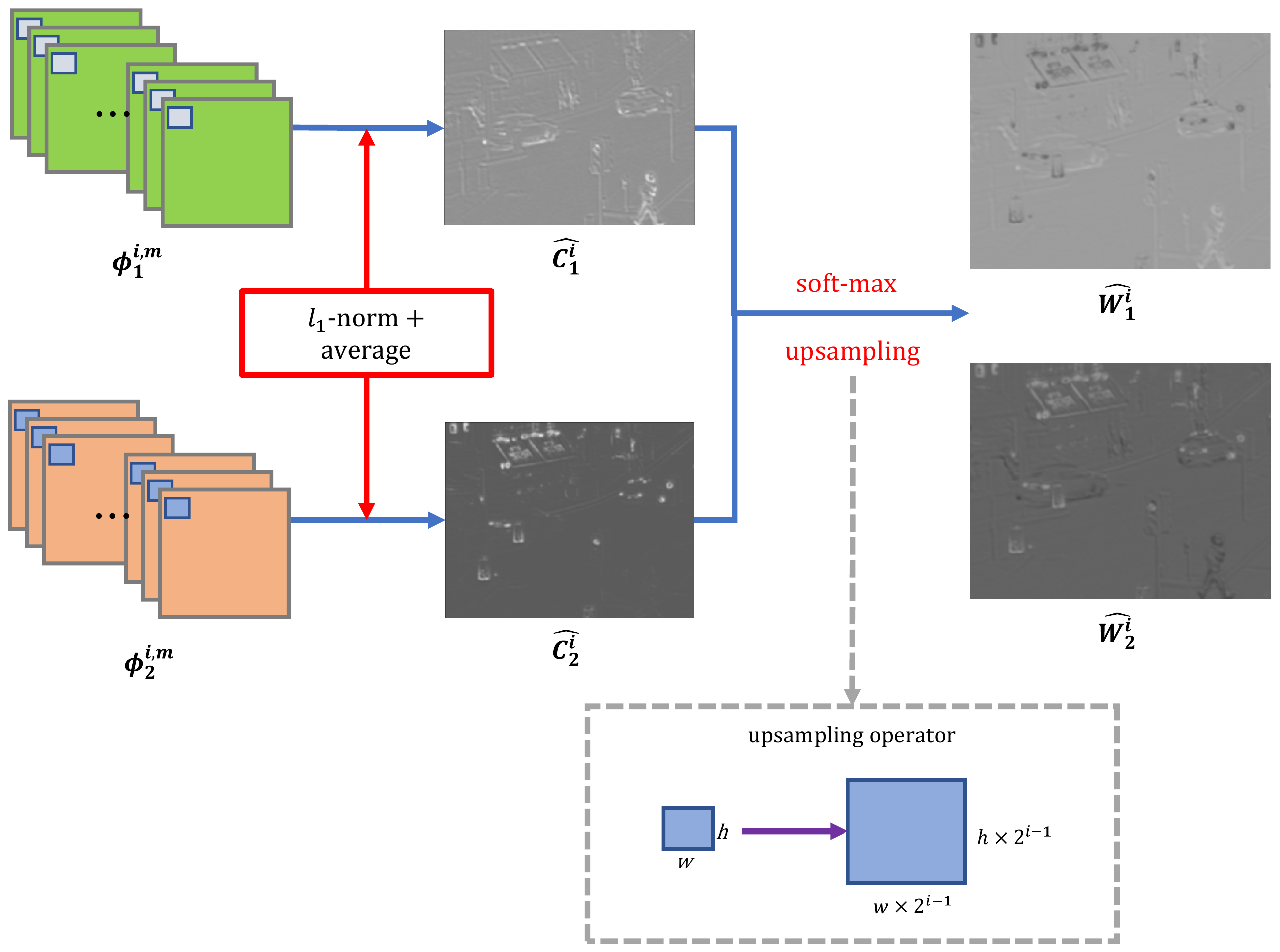}
\caption{The procedure of the fusion strategy for detailed parts.}
\label{fig3}
\end{figure}

Let $\phi_k^{i,1:M}(x,y)$ denote the contents of $\phi_k^{i,m}$ at the position $(x,y)$ in the feature maps. As we can see, $\phi_k^{i,1:M}(x,y)$ is an $M$-dimensional vector. The procedure of our strategy is presented in Fig.\ref{fig3}.

As shown in Fig.\ref{fig3}, after we get the deep features $\phi_k^{i,m}$, the activity level map $\hat{C_k^i}$ will be calculated by $l_1$-norm and block-based average operator, where $k\in{\{1,2\}}$ and $i\in{\{1,2,3,4\}}$.

Inspired by \cite{12}, the $l_1$-norm of $\phi_k^{i,1:M}(x,y)$ can be the activity level measure of the source detail content. Thus, the initial activity level map $C_k^i$ is obtained by
\begin{eqnarray}\label{Eq5}
  	C_k^i(x,y) = ||\phi_k^{i,1:M}(x,y)||_1
\end{eqnarray}

We then use the block-based average operator to calculate the final activity level map $\hat{C_k^i}$ in order to make our fusion method robust to misregistration.
\begin{eqnarray}\label{Eq6}
  	\hat{C_k^i}(x,y) = \frac{\sum_{\beta=-r}^{r}\sum_{\theta=-r}^{r}C_k^i(x+\beta,y+\theta)}{(2r+1)^2}
\end{eqnarray}

\noindent where $r$ determines the block size. The fusion method will be more robust to misregistration if the $r$ is larger, but some detail could be lost. Thus, in our strategy  $r=1$.

Once we get the activity level map $\hat{C_k^i}$, the initial weight maps $W_k^i$ will be calculated by soft-max operator, as shown in Eq.\ref{Eq7},
\begin{eqnarray}\label{Eq7}
  	W_k^i(x,y) = \frac{\hat{C_k^i}(x,y)}{\sum_{n=1}^{K}\hat{C_n^i}(x,y)}
\end{eqnarray}

\noindent where $K$ denotes the number of activity level map, which in our paper is set to $K=2$. $W_k^i(x,y)$ indicates the initial weight map value in the range of [0,1].

As we all know, the pooling operator in VGG-network is a kind of subsampling method. Every time this operator resizes the feature maps to $1/s$ times of the original size where $s$ is the stride of the pooling operator. In the VGG-network, the stride of the pooling operator is 2. So in different layers, the size of feature maps is $1/2^{i-1}$  times the detail content size, where $i\in{\{1,2,3,4\}}$ indicates the layers of $relu\_1\_1$, $relu\_2\_1$, $relu\_3\_1$ and $relu\_4\_1$, respectively. After we get each initial weight map $W_k^i$, we use an upsampling operator to resize the weight map size to the input detail content size.

As shown in Fig.\ref{fig3}, with the upsampling operator, we will get the final weight map $\hat{W_k^i}$, the size of which equals the input detail content size. The final weight map is calculated by Eq.\ref{Eq8},
\begin{eqnarray}\label{Eq8}
  	&\hat{W_k^i}(x+p,y+q) = W_k^i(x,y),\\
	&p,q\in{\{0,1,\cdots,(2^{i-1}-1)\}} \nonumber
\end{eqnarray}

Now we have four pairs of weight maps $\hat{W_k^i}$, $k\in{\{1,2\}}$ and $i\in{\{1,2,3,4\}}$. For each pair $\hat{W_k^i}$, the initial fused detail content is obtained by Eq.\ref{Eq9},
\begin{eqnarray}\label{Eq9}
  	&F_d^i(x,y) = \sum_{n=1}^{K}\hat{W_n^i}(x,y)\times I_n^d(x,y), K=2.
\end{eqnarray}

Finally, the fused detail content $F_d$ is obtained by Eq.\ref{Eq10} in which we choose the maximum value from the four initial fused detail content for each position.
\begin{eqnarray}\label{Eq10}
  	&F_d(x,y) = \max[F_d^i(x,y) | i\in{\{1,2,3,4\}}]
\end{eqnarray}

\subsection{Reconstruction}

Once the fused detail content $F_d$ is obtained, we use the fused base part $F_b$ and the fused detail content $F_d$ to reconstruct the final fused image, as shown in Eq.\ref{Eq11},
\begin{eqnarray}\label{Eq11}
  	&F(x,y) = F_b(x,y)+F_d(x,y)
\end{eqnarray}

\subsection{Summary of the Proposed Fusion Method}

In this section, we summarize the proposed fusion method based on deep learning as follows:
\subsubsection{Image decomposition}
The source images are decomposed by the image decomposition operation\cite{15} to obtain the base part $I_k^b$ and the detail content $I_k^d$, where $k\in{\{1,2\}}$.
\subsubsection{Fusion of base parts}
We choose the weighted-averaging fusion strategy to fuse base parts, with the weight value for each base part of 0.5.
\subsubsection{Fusion of detail content}
The fused detail content is obtained by the multi-layer fusion strategy.
\subsubsection{Reconstruction}
Finally, the fused image is given by Eq.\ref{Eq11}.

\section{Experimental results and analysis}
\label{sec:experiment}
The aim of the experiment is to validate the proposed method using subjective and objective criteria and to carry out a comparison with existing methods.

\subsection{Experimental Settings}

\begin{figure}[!ht]
\centering
\includegraphics[width=\linewidth]{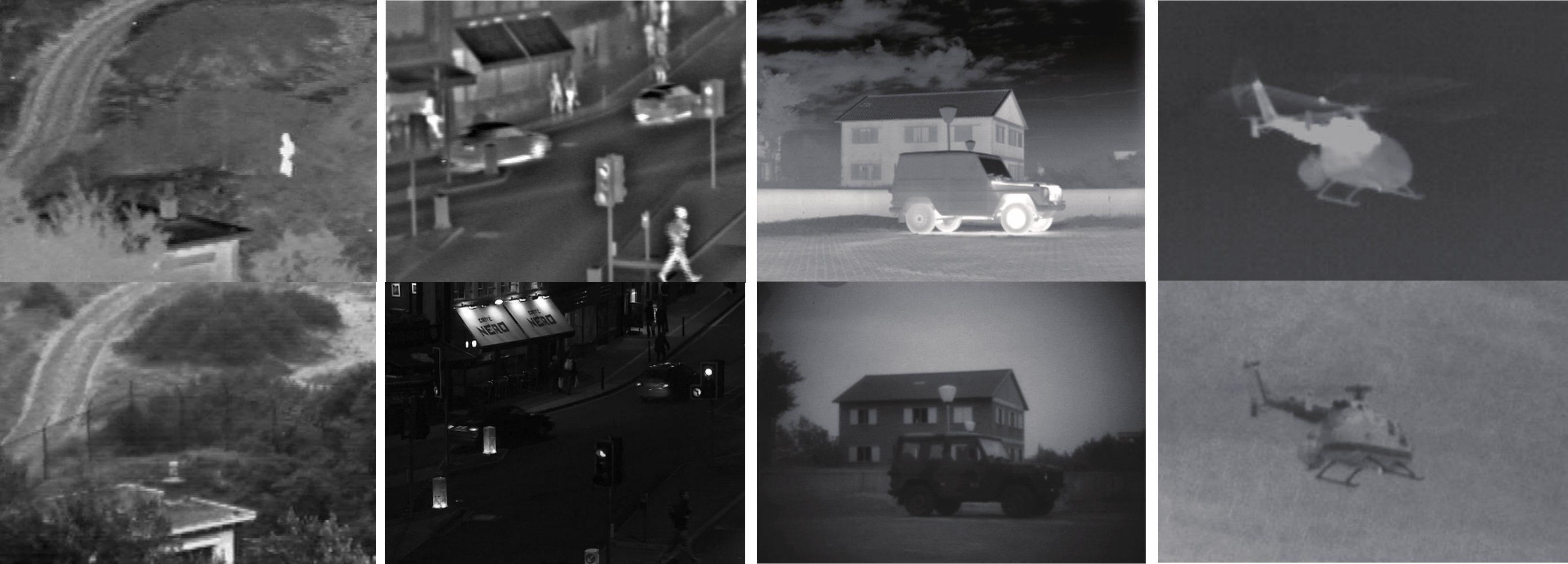}
\caption{Four pairs of source images. The top row contains infrared images, and the second row contains visible images.}
\label{fig4}
\end{figure}

In our experiment, the source infrared and visible images were collected from \cite{20} and \cite{25}. There are 21 pairs of our source images and they are available at \cite{26}. A sample of these images is shown in Fig.\ref{fig4}.

In multi-layer fusion strategy, we choose few layers from a pre-trained VGG-19 network\cite{17} to extract deep features. These layers are $relu\_1\_1$, $relu\_2\_1$, $relu\_3\_1$ and $relu\_4\_1$, respectively

For comparison, we selected several recent and classical fusion methods to perform the same experiment, including: cross bilateral filter fusion method(CBF)\cite{21}, the joint-sparse representation model(JSR)\cite{10}, the JSR model with saliency detection fusion method(JSRSD)\cite{22}, weighted least square optimization-based method(WLS)\cite{20} and the convolutional sparse representation model(ConvSR)\cite{13}.

All the fusion algorithms are implemented in MATLAB R2016a on 3.2 GHz Intel(R) Core(TM) CPU with 12 GB RAM.

\subsection{Subjective Evaluation}

\begin{figure}[ht]
\centering
\includegraphics[width=\linewidth]{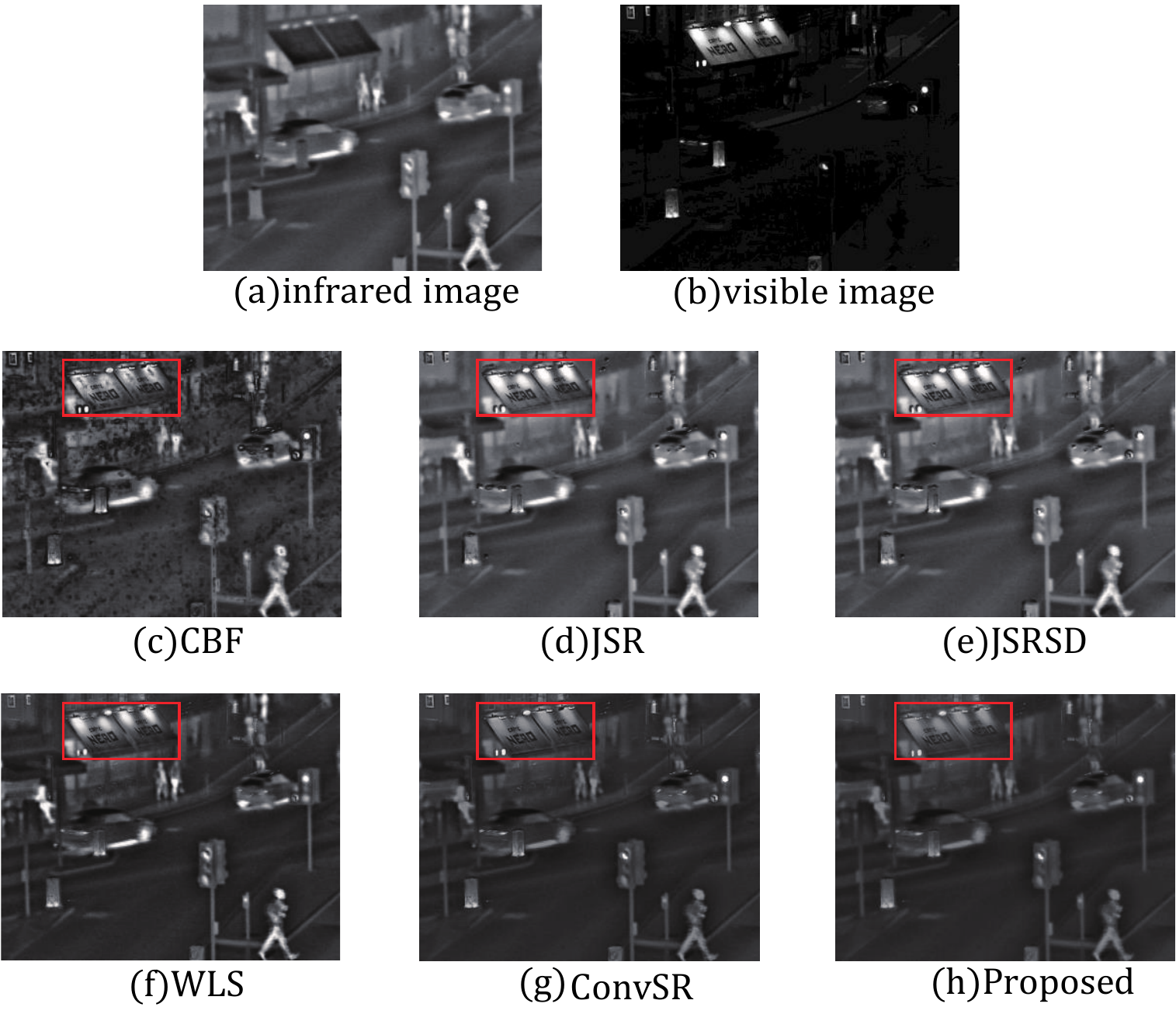}
\caption{Results on ``street'' images. (a) Infrared image; (b) Visible image; (c) CBF; (d) JSR; (e) JSRSD; (f) WLS. (g) ConvSR; (h) The proposed method.}
\label{fig5}
\end{figure}

\begin{figure}[ht]
\centering
\includegraphics[width=\linewidth]{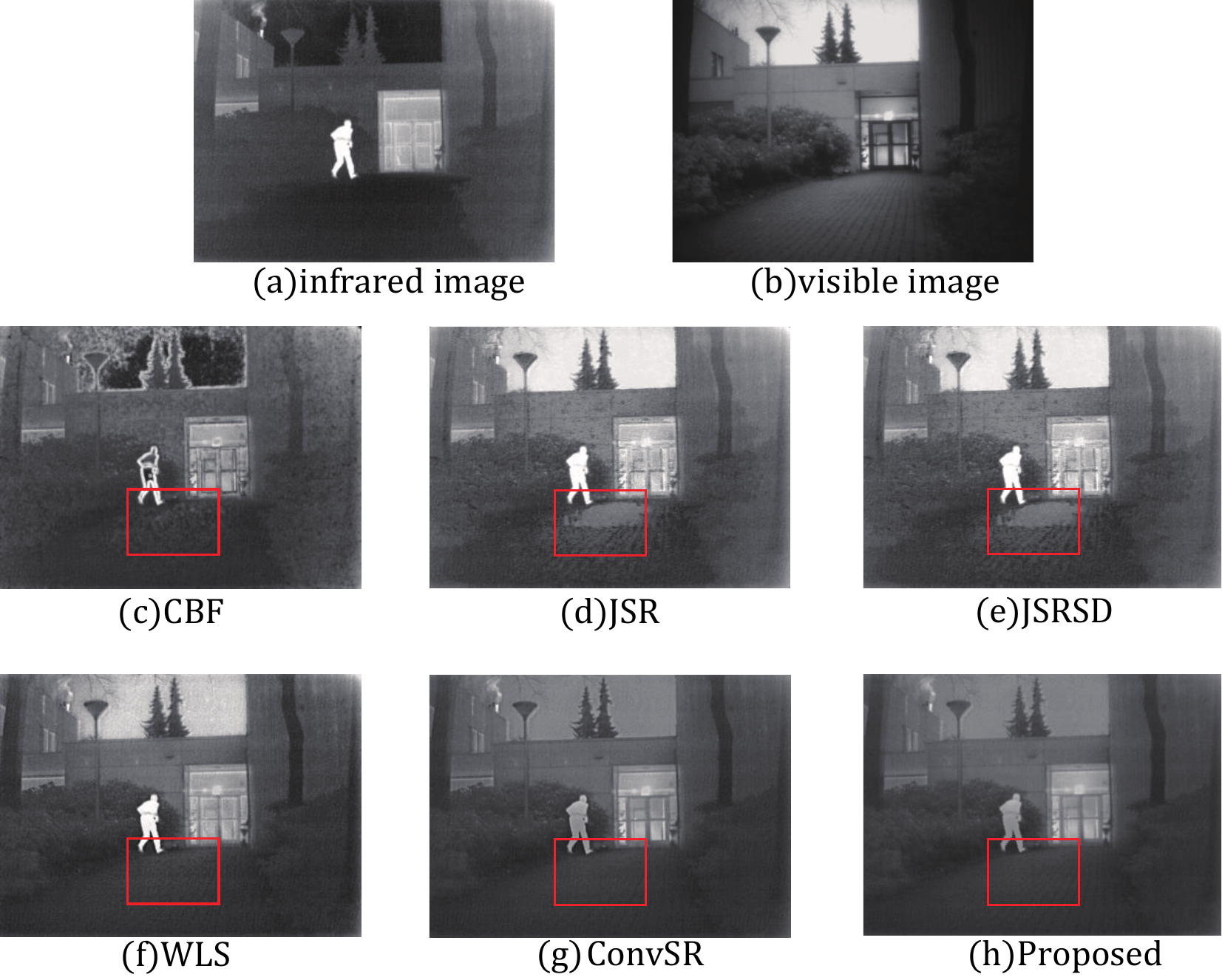}
\caption{Results on ``people'' images. (a) Infrared image; (b) Visible image; (c) CBF; (d) JSR; (e) JSRSD; (f) WLS. (g) ConvSR; (h) The proposed method.}
\label{fig6}
\end{figure}

The fused images which are obtained by the five existing methods and the proposed method are shown in Fig.\ref{fig5} and Fig.\ref{fig6}. Due to the space limit, we evaluate the relative performance of the fusion methods only on a single pair of images(``street'' and``people'').

As we can see from Fig.\ref{fig5}(c-h), the fused image obtained by the proposed method preserves more detail information in the red window and contains less artificial noise. In Fig.\ref{fig6}(c-h), the fused image obtained by the proposed method also contains less noise in the red box.

In summary, the fused images which are obtained by CBF have more artificial noise and the salient features are not clear. The fused images obtained by JSR, JSRSD and WLS, in addition, contain artificial structures around the salient features and the image detail is blurred. In contrast, the fused images obtained by ConvSR and the proposed fusion method contain more salient features and preserve more detail information. Compared with the four existing fusion methods, the fused images obtained by the proposed method look more natural. As there is no visible difference between ConvSR and the proposed fusion method in terms of human sensitivity, we use several objective quality metrics to evaluate the fusion performance in the next section.

\subsection{Objective Evaluation}

For the purpose of quantitative comparison between the proposed method and existing fusion methods, four quality metrics are utilized. These are: $FMI_{dct}$ and $FMI_w$\cite{23} which calculate mutual information (FMI) for the discrete cosine and wavelet features, respectively; $N_{abf}$\cite{24} which denotes the rate of noise or artifacts added to the fused image by the fusion process; and modified structural similarity($SSIM_a$). 

In our paper, the $SSIM_a$ is calculated by Eq.\ref{Eq12},
\begin{eqnarray}\label{Eq12}
  	SSIM_a(F) = (SSIM(F,I_1)+SSIM(F,I_2))\times 0.5
\end{eqnarray}

\noindent where $SSIM(\cdot)$ represents the structural similarity operation, $F$ is the fused image, and $I_1$, $I_2$ are the source images. The value of $SSIM_a$ assesses the ability to preserve structural information.

The performance improves with the increasing numerical index of $FMI_{dct}$, $FMI_{w}$ and $SSIM_a$. On the contrary, the fusion performance is better when the value of $N_{abf}$ is small, which means the fused images contain less artificial information and noise.

The average values of $FMI_{dct}$, $FMI_{w}$, $SSIM_a$ and $N_{abf}$ obtained by teh existing methods and the proposed method for the 21 fused images are shown in Table \ref{Table1}.
\begin{table}[ht]
\scriptsize
\centering
\caption{\label{Table1}The average values of $FMI_{dct}$, $FMI_{w}$, $SSIM_a$ and $N_{abf}$ for 21 fused images.}
\resizebox{3.4in}{!}{
\begin{tabular}{|c|c|c|c|c|c|c|}
\hline
Methods  		& CBF\cite{21}		&	WLS\cite{20}		&	JSR\cite{10}		&	JSRSD\cite{22}	&	ConvSR\cite{13}	&	Proposed \\
\hline
$FMI_{dct}\cite{23}$	&	0.26309	&	0.33103	&	0.14236	&	0.14253	&	0.34640	&	\textbf{0.40463} \\
\hline
$FMI_{w}\cite{23}$		&	0.32350	&	0.37662	&	0.18506	&	0.18498	&	0.34640	&	\textbf{0.41684} \\
\hline
$SSIM_{a}$	&	0.59957	&	0.72360	&	0.54073	&	0.54127	&	0.75335	&	\textbf{0.77799} \\
\hline
$N_{abf}\cite{24}$		&	0.31727	&	0.21257	&	0.34712	&	0.34657	&	0.0196	&	\textbf{0.00120} \\
\hline
\end{tabular}}
\end{table}

In Table \ref{Table1}, the best values for $FMI_{dct}$, $FMI_{w}$, $SSIM_a$ and $N_{abf}$ are indicated in bold. As we can see, the proposed method has all the best average values for these metrics. These values indicate that the fused images obtained by the proposed method are more natural and contain less artificial noise. From the objective evaluation, our fusion method has better fusion performance than the existing methods.

\begin{figure}[ht]
\centering
\includegraphics[width=\linewidth]{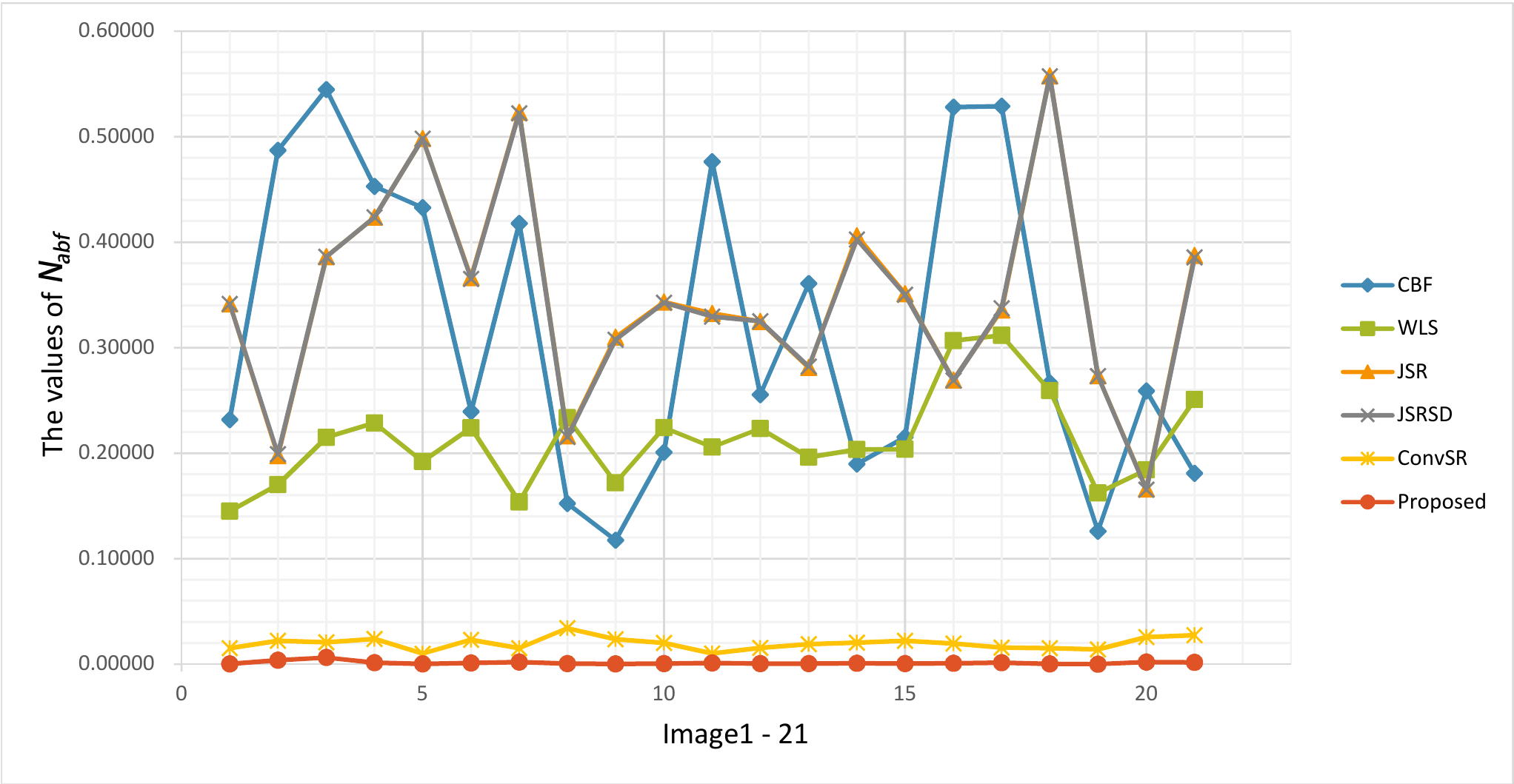}
\caption{Plotting $N_{abf}$ for all fused images obtained by the fusion methods experimentally compared.}
\label{fig7}
\end{figure}

Specifically, we show in Table \ref{Table2} all values of $N_{abf}$ for the 21 pairs produced by the respective methods. The graph plot of $N_{abf}$ for all fused images is shown in Fig.\ref{fig7}.

\begin{table}[ht]
\scriptsize
\centering
\caption{\label{Table2}The $N_{abf}$ values for 21 fused images which obtained by fusion methods.}
\resizebox{3.4in}{!}{
\begin{tabular}{|c|c|c|c|c|c|c|}
\hline
Methods  	& CBF\cite{21}	&	WLS\cite{20}	&	JSR\cite{10}	&	JSRSD\cite{22}	&	ConvSR\cite{13}	&	Proposed \\
\hline
image1	& 0.23167	& 0.14494	& 0.34153	& 0.34153	& 0.0149	& \textbf{0.00013} \\
\hline
image2	& 0.48700	& 0.16997	& 0.19749	& 0.19889	& 0.0220	& \textbf{0.00376} \\
\hline
image3	& 0.54477	& 0.21469	& 0.38627	& 0.38627	& 0.0207	& \textbf{0.00622} \\
\hline
image4	& 0.45288	& 0.22866	& 0.42353	& 0.42353	& 0.0238	& \textbf{0.00132} \\
\hline
image5	& 0.43257	& 0.19188	& 0.49804	& 0.49804	& 0.0099	& \textbf{0.00020} \\
\hline
image6	& 0.23932	& 0.22382	& 0.36619	& 0.36509	& 0.0230	& \textbf{0.00099} \\
\hline
image7	& 0.41779	& 0.15368	& 0.52301	& 0.52220	& 0.0151	& \textbf{0.00188} \\
\hline
image8	& 0.15233	& 0.23343	& 0.21640	& 0.21536	& 0.0340	& \textbf{0.00037} \\
\hline
image9	& 0.11741	& 0.17177	& 0.30983	& 0.30761	& 0.0237	& \textbf{0.00029} \\
\hline
image10	& 0.20090	& 0.22419	& 0.34329	& 0.34271	& 0.0201	& \textbf{0.00048} \\
\hline
image11	& 0.47632	& 0.20588	& 0.33225	& 0.32941	& 0.0102	& \textbf{0.00109} \\
\hline
image12	& 0.25544	& 0.22335	& 0.32488	& 0.32502	& 0.0154	& \textbf{0.00058} \\
\hline
image13	& 0.36066	& 0.19607	& 0.28106	& 0.28220	& 0.0189	& \textbf{0.00035} \\
\hline
image14	& 0.18971	& 0.20332	& 0.40615	& 0.40261	& 0.0204	& \textbf{0.00082} \\
\hline
image15	& 0.21509	& 0.20378	& 0.35106	& 0.35013	& 0.0221	& \textbf{0.00060} \\
\hline
image16	& 0.52783	& 0.30672	& 0.26907	& 0.26888	& 0.0194	& \textbf{0.00090} \\
\hline
image17	& 0.52887	& 0.31160	& 0.33544	& 0.33720	& 0.0156	& \textbf{0.00122} \\
\hline
image18	& 0.26649	& 0.25937	& 0.55761	& 0.55732	& 0.0150	& \textbf{0.00023} \\
\hline
image19	& 0.12582	& 0.16205	& 0.27327	& 0.27302	& 0.0138	& \textbf{0.00002} \\
\hline
image20	& 0.25892	& 0.18401	& 0.16588	& 0.16541	& 0.0257	& \textbf{0.00203} \\
\hline
image21	& 0.18091	& 0.25074	& 0.38734	& 0.38546	& 0.0275	& \textbf{0.00171} \\
\hline
\end{tabular}}
\end{table}

From Table \ref{Table2} and Fig.\ref{fig7}, the values of $N_{abf}$ produced by our method are nearly two orders of magnitude batter than CBF, JSR and JSRSD. Even compared with ConvSR, the $N_{abf}$ values of proposed method are extremely small. This indicates that the fused images obtained by the proposed method contain less artificial information and noise.

\section{Conclusion}
\label{sec:con}
In this paper, we present a simple and effective fusion method based on a deep learning framework(VGG-network) for an infrared and visible image fusion task. Firstly, the source images are decomposed into base parts and detail content. The former contains low frequency information and the latter contains texture information. These base parts are fused by the weight-averaging strategy. For the detail content, we proposed a novel multi-layer fusion strategy based on a pre-trained VGG-19 network. The deep features of the detail content are obtained by this fixed VGG-19 network. The $l_1$-norm and block-averaging operator are used to get the initial weight maps. The final weight maps are obtained by the soft-max operator. The initial fused detail content is generated for each pair of weight maps and the input detail content. The fused detail content is reconstructed by the max selection operator applied to these initial fused detail content. Finally, the fused image is reconstructed by adding the fused base part and the fused detail content. We use both subjective and objective methods to evaluate the proposed method. The experimental results show that the proposed method exhibits state-of-the-art fusion performance.

We believe our fusion method and the novel multi-layer fusion strategy can be applied to other image fusion tasks, such as medical image fusion, multi-exposure image fusion and multi-focus image fusion.

\end{document}